\pdfoutput=1
\documentclass[11pt]{article}

\usepackage[final]{acl}

\usepackage{times}
\usepackage{latexsym}

\usepackage{multirow}
\usepackage{array}
\usepackage{tabularx}
\usepackage{adjustbox}
\usepackage{graphicx}
\usepackage{amsmath}
\usepackage{caption}
\usepackage{subcaption}
\usepackage{makecell}
\usepackage{hyperref}
\usepackage{bm}
\usepackage{xcolor}
\usepackage{colortbl}
\usepackage[T1]{fontenc}
\usepackage[utf8]{inputenc}

\usepackage{microtype}

\usepackage{inconsolata}

\title{Enhancing Idiomatic Representation in Multiple Languages via an Adaptive Contrastive Triplet Loss}

\author{Wei He \\
  University of Sheffield \\
  \texttt{wei.he1@sheffield.ac.uk} \\\And
  Marco Idiart \\
  Federal University of Rio Grande do Sul \\
  \texttt{marco.idiart@gmail.com} \\\AND
  Carolina Scarton \\
  University of Sheffield \\
  \texttt{c.scarton@sheffield.ac.uk } \\\And
  Aline Villavicencio \\
  University of Exeter and University of Sheffield \\
  \texttt{A.Villavicencio@exeter.ac.uk } 
  }

\begin{document}
\maketitle
\begin{abstract}

Accurately modeling idiomatic or non-compositional language has been a longstanding challenge in Natural Language Processing (NLP). 
This is partly because these expressions do not derive their meanings solely from their constituent words, but also due to the scarcity of relevant data resources, and their 
impact on the performance of downstream tasks such as machine translation and simplification.  In this paper we propose an approach to model idiomaticity effectively using a triplet loss 
that incorporates the asymmetric contribution of components words to an idiomatic meaning for training language models by using adaptive contrastive learning and resampling miners to build an idiomatic-aware learning objective. Our proposed method is evaluated on a SemEval challenge and outperforms previous alternatives significantly in many metrics. Our code is available at our project\footnote{https://github.com/risehnhew/Enhancing-Idiomatic-Representation-in-Multiple-Languages}.
\end{abstract}

\section{Introduction }
Among multiword expressions (MWEs), idiomatic expressions (IEs) are difficult to model as their meaning is often not straightforwardly related to the meaning of the component words \cite{sag2002multiword}. These expressions, which are also commonly referred to as non-compositional expressions, often take on figurative meanings. For example, \textit{eager beaver} has a figurative meaning of \textit{an enthusiastic person who works very hard} different from the literal meanings of its component words like \textit{impatient rodent} \citep{sag2002multiword, villavicencio2019discovering}. They are a common occurrence across various genres \citep{haagsma-etal-2020-magpie}.   

Accurately understanding idiomatic expressions has posed a significant challenge, as word and phrase representations may favor inherently compositional usages at the levels of both words and sub-words to minimize their vocabulary \citep{gow-smith-etal-2022-improving}. 
%Modern Natural Language Processing  systems 
Indeed recent models are mainly driven by compositionality, which is at the core of tokenization \citep{sennrich-etal-2016-neural} and self-attention mechanism \citep{vaswani2017attention}. 
%According to a recent study 
Pre-trained language models including static and contextualised embeddings do not seem to be well-equipped to capture the meanings of IEs, as %the meanings of IE and their embeddings are demonstrated by the fact that 
IEs with similar meanings are not close in the embedding space \citep{garcia-etal-2021-probing}. This reveals a need for models that can accurately capture idiomatic language.  
Ensuring precise representation of IEs is crucial for their precise handling in various downstream applications, such as sentiment analysis \citep{liu-etal-2017-idiom, biddle2020leveraging}, dialog models \citep{jhamtani-etal-2021-investigating} and text simplification \citep{he2023representation}. 

To address this issue, previous methods often rely on new datasets with human annotations or on data augmentation \citep{liu2023crossing, dankers-lucas-2023-non}. However,  
%using existing data resources may also work with 
the use of alternative training processes has also been effective, including  regression objective functions with a siamese network \citep{tayyar-madabushi-etal-2021-astitchinlanguagemodels-dataset} or substitute objectives \citep{liu-etal-2022-ynu} to break the compositionality of idiomatic phrases, as finding an objective to stand for idiomatic representation is difficult.

% In this work, we take the definition of idiomatic-aware models in \citet{tayyar-madabushi-etal-2022-semeval} task 2, a dedicated designed triplet loss adapted to the task as the objective function to fine-tuning a pre-trained model by using in-batch positive anchor negative triplets \citep{balntas2016learning} for building an idiomatic-aware language model. Idiomatic-aware models are trained on triplet text, where the positive and anchor are sentences with idiomatic expressions and synonyms interchangeably. The "learn-to-compare" paradigm of contrastive learning has become a widely used framework to obtain better text embeddings\citep{ni-etal-2022-large,ni-etal-2022-sentence,wang2022text}, which fits for distinguishing figurative from literal meanings of MWEs.
Our work focuses on the development of idiomatic-aware language models, which are designed to better represent MWEs of various degrees of idiomaticity in natural language text. To achieve this, we adopt the definition of idiomatic-aware models from SemEval 2022 task 2 \citep{tayyar-madabushi-etal-2022-semeval} that when using the model, the semantic similarity between an IE and its incorrect paraphrase equals the semantic similarity between a correct and an incorrect paraphrase. Our approach involves fine-tuning a pre-trained model using a bespoke triplet loss function that is specifically designed for capturing the asymmetry between the surface forms of the component words and their semantic contribution to the meaning of the expression. To build this idiomatic-aware language model, we use in-batch positive-anchor-negative triplets \citep{balntas2016learning}.
\begin{figure}[t] % 
    \centering
    \includegraphics[width=0.4\textwidth]{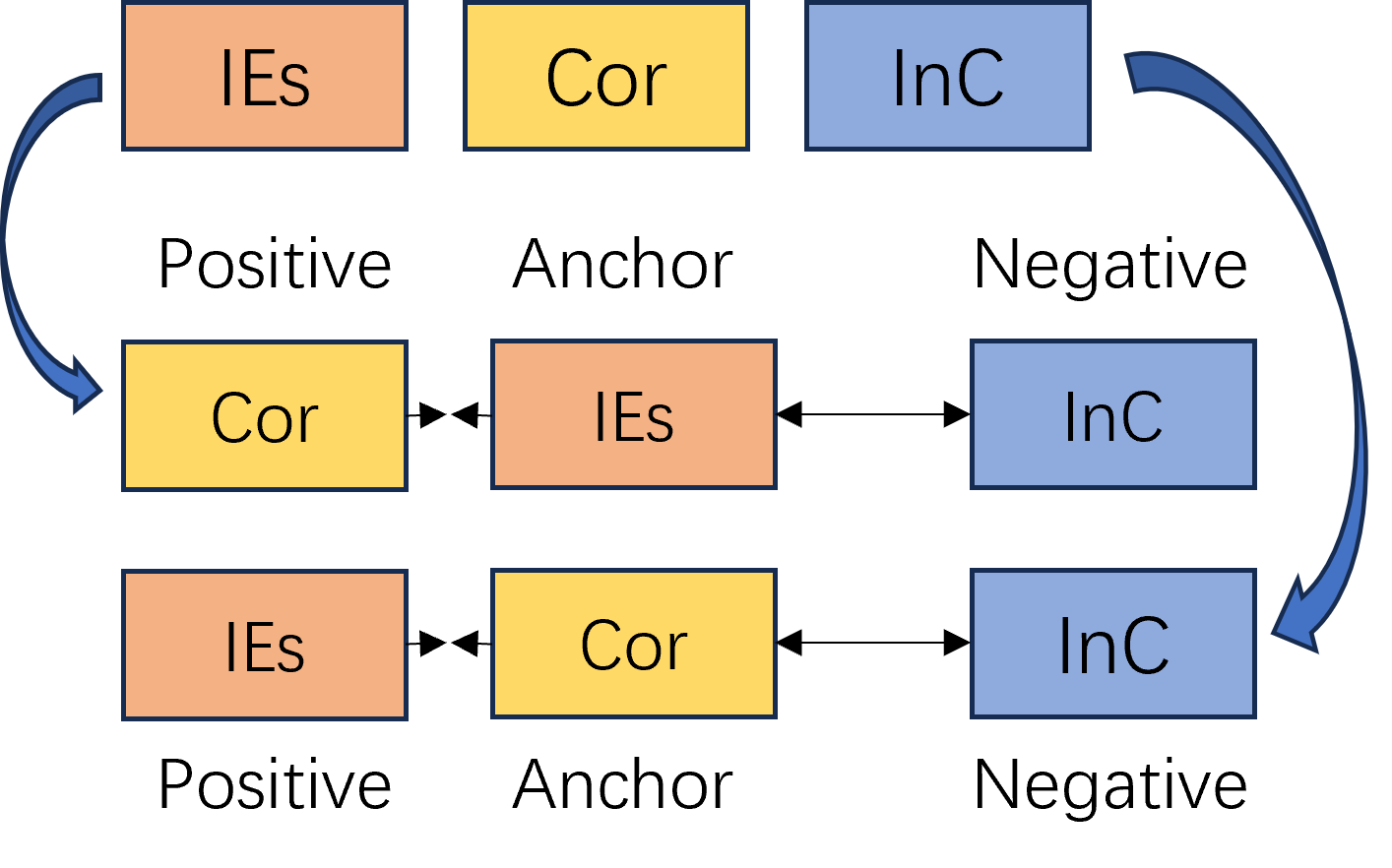} 
    \caption{Triplet Resampling by using a specifically designed miner. For a triplet, it can generate 2 samples by treating the sentence containing IEs (IEs) and a correct (Cor) paraphrase sentence as positive and anchor (and vice-versa) interchangeably and the Incorrect (InC) sentence as a negative sample.}
    \label{fig:triplet}
\end{figure}
Our model is trained on extracted triplets, where sentences with the idiomatic expressions and their synonyms  correspond to positive and anchor respectively, and vice-versa, Figure \ref{fig:triplet}. The aim of this training is to enable the model to learn the difference between the literal meanings of the component words of an MWE when used in isolation and their idiomatic meanings as part of the MWE. We use the "learn-to-compare" paradigm of contrastive learning (CL), which has been successfully adopted for obtaining better text embeddings \citep{ni-etal-2022-large,ni-etal-2022-sentence,wang2022text} including for polysemous words \cite{liu2019single}. This framework fits well with our objective of distinguishing between the figurative and literal meanings of MWEs.

% In summary, our approach involves using a triplet loss function, in-batch positive anchor negative triplets, and the "learn-to-compare" paradigm of contrastive learning to build an idiomatic-aware language model. This model is trained to distinguish between the figurative and literal meanings of MWEs, which will improve its understanding of natural language text.

% A series of based models in different sizes and pre-training strategies are trained in our proposed training loss.
% Our best models achieve new state-of-the-art results with a significant improvement of $11\%$ in idiom only and $2.5\%$ in over-all over the previous best in evaluation task. 
To evaluate the approach, a set of models with varying sizes and pre-training strategies is trained using this novel training method that we proposed. %After extensive experimentation, %we identified 
The best models achieved new state-of-the-art results in the dataset containing expressions of varying levels of idiomaticity, and %Specifically, 
our best model demonstrated a substantial improvements in both %of $21.9\%$ in 
idiom-only performance and  %$3.6\%$ boost 
overall performance compared to the previous best results. 
%This highlights the effectiveness of our proposed approach and its potential to push the boundaries of natural language processing further.
Our contributions are:
\begin{description}
\item[An efficient approach] %We present an simple and efficient approach 
for creating language models 
that can represent MWEs of varying levels of idiomaticity. This is achieved through a specialized training process using a triplet loss function and in-batch positive-anchor-negative triplets.

% Contrastive Learning Framework: Employing the "learn-to-compare" paradigm from contrastive learning, our models are trained to discern the nuanced differences between figurative and literal meanings of MWEs. This framework is key in enhancing the model's capability to process idiomatic language effectively.

% Integration with Pre-Trained Semantic-Aware Models: To overcome the challenges posed by limited idiomatic text data, our models are initialized using existing pre-trained semantic-aware models. This integration provides a robust foundation for our models to build upon, enhancing their learning efficiency and effectiveness.

\item[An idiomatic-aware loss function] 
%We introduce a  training loss function 
tailored to directly optimize the 
%definition of a good understanding 
representation of idiomatic language and the potentially asymmetric and non-compositional contributions of the component words. This function plays a crucial role in training to discern the nuanced differences between idiomatic and literal meanings of MWEs.

% Extensive Experimental Validation: Through rigorous experimentation, we have validated the effectiveness of our approach. Our models have shown remarkable improvements in idiom-only performance, with an $11\%$ increase, and a significant $2.5\%$ overall performance boost compared to previous benchmarks. These results underscore the strength of our methods and their potential to significantly advance the field of natural language processing.

\item[New state-of-the-art performance] 
%The culmination of our efforts has led to the creation of 
models %that set a new standard in the processing and 
that enable the understanding of idiomatic language. 
%within natural language processing systems. 
This advancement represents a major leap forward 
%in the field, 
opening up new possibilities for more nuanced and accurate language understanding.
%in various applications.
\end{description}

The paper starts with an overview of %existing challenges and solutions of i
previous work on idiomaticity representation in Section \ref{sec:re}. It also introduces contrastive learning in NLP and IE evaluation methods. Section \ref{sec:Id} presents our method using a triplet loss and data mining to do efficient training. Section \ref{sec:Ex} describes our experiments, and Section \ref{sec:Re} analyzes the results. 

\section{Related Work}
\label{sec:re}
%\subsection{Idiomaticity Representation Method}

Idiomaticity representation can be challenging 
even for large language models \citep{king-cook-2018-leveraging, nandakumar-etal-2019-well,cordeiro-etal-2019-unsupervised,hashempour-villavicencio-2020-leveraging, garcia-etal-2021-probing, klubicka-etal-2023-idioms}. For instance,  GPT-3 \citep{brown2020language} reaches only $50.7\%$ accuracy in idiom comprehension \citep{zeng2022getting}. This may be possibly due to idiomatic expressions being non-compositional and having figurative meanings that go beyond its individual words \citep{baldwin2010multiword}.
%%AV from original: 
%% for natural language processing, as idiomatic expressions are often non-compositional and have figurative meanings \citep{baldwin2010multiword}. Many studies have tested that even the recent large language models cannot represent IEs well \citep{king-cook-2018-leveraging, nandakumar-etal-2019-well,cordeiro-etal-2019-unsupervised,hashempour-villavicencio-2020-leveraging, garcia-etal-2021-probing, klubicka-etal-2023-idioms}.  \citet{zeng2022getting} tested the GPT-3 model for its idiom comprehension and found that only $50.7\%$ accuracy can be acquired by this flagship model. 

Methods that have been used for representing idiomaticity include combining compositional components with adaptive weights \citep{hashimoto-tsuruoka-2016-adaptive, li2018adaptive}, representing MWEs with single tokens \citep{yin-schutze-2015-discriminative, li2018phrase,cordeiro2019unsupervised,phelps2022drsphelps} and creating phrase embeddings that effectively capture both compositional and idiomatic %(non-compositional) 
expressions \citep{hashimoto-tsuruoka-2016-adaptive}. The latter involves an adaptive learning process that adjusts to the nature of the phrases to generate accurate representation.
An adapter-based approach is proposed that augmenting the BART model with an "idiomatic adapter" trained on dedicated idiom datasets \citep{zeng-bhat-2022-getting}. This adapter acts as a lightweight expert, enhancing BART's ability to capture figurative meanings alongside literal interpretations. %A later study propose 
PIER \citep{zeng-bhat-2023-unified}, a language model based on BART,  specifically addresses the challenge of representing non-compositional expressions, such as idioms, in natural language. Traditional compositionality-based models often struggle with these expressions, as their meaning cannot be simply derived from the sum of their parts. PIER overcomes this by incorporating an "idiomatic adapter" which learns to represent figurative meanings alongside literal ones.
Additionally, \citet{liu-etal-2023-crossing} proposed a novel approach to idiomatic machine translation through retrieval augmentation and loss weighting, which significantly improves the translation quality of idiomatic expressions by leveraging context retrieval mechanisms and adjusting loss functions to better handle idiomaticity.

\paragraph{Contrastive Learning}

Contrastive learning is a method in machine learning that trains a model to distinguish between similar and dissimilar pairs of data.
 In recent times, significant progress has been made in sentence embeddings through contrastive learning \citep{gao-etal-2021-simcse, giorgi-etal-2021-declutr, kim-etal-2021-self, wu-etal-2022-pcl, zhang2022unsupervised, xu-etal-2023-simcse}. 
It also has been widely applied in other NLP research fields, such as text classification \citep{fang2020cert}, machine translation \citep{pan-etal-2021-contrastive}, information extraction \citep{qin-etal-2021-erica}, question answering \citep{karpukhin-etal-2020-dense} and text retrieval \citep{xiong2020approximate}. 
Despite their shared goal of acquiring high-quality text representations \citep{reimers-gurevych-2019-sentence, gao-etal-2021-simcse,neelakantan2022text,giorgi2021declutr}, the exploration of idiomatic representation and related research through contrastive learning is still yet to be fully explored. 
Contrastive learning with triplet loss involves training the model on triplets: an anchor sample, a positive sample (similar to the anchor), and a negative sample (dissimilar to the anchor). The goal is to minimize the distance between anchors and positive samples while maximizing the distance between anchors and negative samples. This approach has recently been applied to tasks such as idiom usage recognition and metaphor detection \citep{zhou-etal-2023-clcl}. 
% Different from this method, our method anchor and positive to sentences with IE and positive example
\paragraph{Idiomaticity Representation Evaluation }
% To evaluate the idiomaticity representation in multilingual text, SemEval-2022 proposed task 2B, a novel task that requires models to predict the semantic text similarity (STS) scores between sentence pairs, regardless of whether or not either sentence contains an idiomatic expression. 

% The task was designed to address the shortcomings of existing state-of-the-art models, which often struggle to handle idiomaticity. The results of the task show that it is possible to develop models that can reliably detect and represent idiomaticity in multiple languages.

% The task is also significant because it is the first shared task to focus on both idiomaticity detection and representation. This is important because idiomaticity is a complex phenomenon that can be manifested in different ways. For example, an idiomatic expression may have a non-literal meaning, or it may have a different meaning depending on the context in which it is used.

% The results of the task will be useful for researchers and developers working on natural language processing (NLP) systems that need to handle idiomaticity. For example, machine translation systems need to be able to translate idiomatic expressions correctly, and question answering systems need to be able to understand and answer questions that contain idiomatic expressions.
%Evaluating idiomaticity representation for a language model is a complex task, as capturing the nuances of non-literal expressions goes beyond simply understanding individual words.
%It commonly has intrinsic and extrinsic evaluations. 

Assessing idiomatic representation in language models has  included both  extrinsic and intrinsic  evaluations.
Extrinsic methods evaluate how well the model's idiomaticity representation impacts downstream tasks, such as machine translation \citep{dankers-etal-2022-transformer}, sentence generation \citep{zhou-etal-2021-pie} or conversational systems \citep{adewumi2022vector}.
% The outcome of the task will be useful for researchers and developers working on NLP systems that need to handle idiomaticity. For instance, machine translation systems need to be able to translate idiomatic expressions correctly, and question answering systems need to be able to understand and answer questions that contain idiomatic expressions.
Intrinsic methods evaluate the model's understanding of idiomaticity itself, using approaches like probing to investigate and understand the linguistic information encoded in the representation \citep{garcia-etal-2021-assessing}. Datasets like AStitchInLanguageModels \citep{tayyar-madabushi-etal-2021-astitchinlanguagemodels-dataset} and Noun Compound Type and Token Idiomaticity
(NCTTI) dataset \citep{garcia-etal-2021-assessing} offer labelled examples for intrinsically testing how much the similarities perceived by a model are compatible with human judgements about similarity.
% The NCTTI  is a dataset of NCs
% in English and Portuguese annotated at type and
% token level with human judgments about idiomaticity,
% and with suggestions of paraphrases.  It consists of 8,725 annotations and 5,091 annotations in English and Portuguese, respectively. It evaluates the effectiveness of vector space models in maintaining the meaning of Non-Compositional Phrases (NCs) in different contexts and with lexical substitutions.  This insight can help improve the accuracy of contextualized models in both controlled and naturalistic conditions in English and Portuguese \citep{garcia-etal-2021-assessing}.
More broadly, SemEval-2022  task 2B \citep{tayyar-madabushi-etal-2022-semeval},  evaluates idiomaticity representation in multilingual text while also requiring models to predict the semantic text similarity (STS) scores between sentence pairs, regardless of whether or not either sentence contains an idiomatic expression. The main objective of this task is to address the shortcomings of existing state-of-the-art models, which often struggle to handle idiomaticity. We use this dataset to evaluate our methods.

\section{Idiomaticity-aware Objective}
\label{sec:Id}
Our strategy for improving IE representation in language models utilizes a contrastive triplet loss  adapted  to prioritize idiomaticity and employs a miner to generate positive-anchor-negative triplets for training the model.

\subsection{Triplet Loss}
Triplet loss is a powerful tool for training language models to learn representations of data that are useful for a variety of NLP tasks \citep{neculoiu-etal-2016-learning}. 
% [IT SEEMS CONTRADICTORY AND NOT THE RIGHT PLACE] It is particularly well-suited for tasks where the data is not easily labeled with triplet extraction techniques \citep{xu-etal-2021-learning}. 
It has also been widely used in training models for tasks such as image retrieval, and face recognition \citep{schroff2015facenet, khosla2020supervised}.

Triplet loss is a distance-based loss function defined as
\begin{align} \label{eq3}
   L_{a, p ,n} = max(d(a_i, p_i)-d(a_i, n_i) + m, 0 ),
\end{align} 
where the triplets  $(a_{i},p_{i},n_{i}), i=1\cdots N$, correspond to  $anchor$, $positive$ and $negative$ examples, where  $a_{i}$ and $p_{i}$ are semantically identical and $n_{i}$ is semantically dissimilar from them. $d(x, y)$ is a distance measure and in our method we use cosine similarity (denoted here by \emph{sim})
\begin{equation}
d(x,y) = sim(x,y).
\end{equation}
Finally, the margin \emph{m} controls the minimum distance between anchor-positive pairs and anchor-negative pairs.

%The deep learning model is used to extract features from the input data, and the triplet loss function is then used to train the model to learn embeddings that are closer for similar data points and farther for dissimilar data points. It is a distance-based loss function that aims to learn embeddings that are closer for similar input data and farther for dissimilar ones. 

%Triplets are defined as follows: $\{(a_{i}, p_{i}, n_{i})\}^{N}_{i=1}$, where $(a_{i},p_{i},n_{i})$ corresponds to a triplet with an $anchor$, $positive$ and $negative$, indicating that $a_{i}$ and $p_{i}$ are semantically identical and $n_{i}$ is semantically dissimilar from them. 

% Triplet loss takes inputs with three elements: an $anchor$, a $positive$ example, and a $negative$ example. 

%The goal of triplet loss is to minimize the distance between the $anchor$ input and the $positive$ example, while maximizing the distance between the $anchor$ input and the $negative$ example. It helps the model distinguish between similar and dissimilar items. This forces the model to learn embeddings that are closer for similar data points and farther for dissimilar data points. The anchor input is the reference point that the model is trying to learn an embedding for. The positive example is a similar data point, and the negative example is a dissimilar data point.

Selecting the right margin is crucial for our method. If it is too small, the task becomes too easy, lacking meaningful distinctions. Conversely, if it is too large, it can slow down convergence or yield suboptimal solutions \citep{schroff2015facenet}. The margin is a hyperparameter and its tuning requires experimentation based on the specific dataset and application.

%It is important to choose the appropriate value for the margin in our method. A margin that is too small might make the optimization problem too easy, but it won't ensure meaningful separations. Conversely, a margin that is too large might make the optimization problem too hard, slowing down convergence or leading to suboptimal solutions \citep{schroff2015facenet}. The  margin is an hyperparameter and its tuning requires experimentation based on the specific dataset and application.

%In our method, we use cosine similarity ($sim$) to calculate the distance. 
% A triplet $(a, p, n)$ is a set of three samples used in language modeling systems to determine if two sentences have the same meaning or not. 

% [THE PARAGRAPH BELOW SEEMS TO BE RELATED TO TRIPLET SELECTION. SHOULD IT BE MOVED?]

In this paper we  use a miner to build triplets for learning idiomaticity more efficiently.

% Given a triplet of $(a, p, n)$, it is valid only if:

% $a$ and $p$ has the same label,
% $a$ and $p$ are distinct samples,
% and $n$ has a different label from $a$ and $p$.

\subsection{Modelling IEs with Adaptive Contrastive Tripet Loss}
This section explains how to improve the language model's ability to understand IEs in text without STS scores by adapting triplet loss to the IE-aware training strategy. We will describe the process step-by-step and discuss its benefits.

\subsubsection{Task Definition}
One widely used approach for measuring idiomaticity is by calculating the distance between a dedicated representation for the MWE as a single token and a compositional representation of its components using operations like sum or multiplication \citep{mitchell-lapata-2008-vector,cordeiro2019unsupervised}.
A good idiomatic expression representation,as framed by \citet{madabushi2022semeval}, should have the following property:
\begin{equation}
\begin{aligned}
 \label{eq0}
sim(S_{MWE}, S_{\rightarrow c} ) &= 1 \\
sim(S_{MWE},S_{\rightarrow i}) &= sim(S_{\rightarrow c}  ,S_{\rightarrow i})
\end{aligned}
\end{equation}
where $S_{MWE}$ denotes a sentence containing the idiomatic expression and $S_{\rightarrow c}$ and $S_{\rightarrow i}$ represent sentences with the idiomatic expression replaced by its correct and incorrect paraphrases, respectively. Ensuring these properties hold for all MWEs during training using standard loss functions can be challenging.

Previous studies need annotated similarity scores of pairs as labels for building the training set \citep{tayyar-madabushi-etal-2021-astitchinlanguagemodels-dataset,phelps2022drsphelps}. Their objective functions are as follows:
\begin{equation}
\begin{aligned}
 \label{eq2}
sim(S_{MWE}, S_{\rightarrow c} ) &= 1 \\
sim(S_{MWE},S_{\rightarrow i}) &= score_1 \\
sim(S_{\rightarrow c}  ,S_{\rightarrow i}) &= score_2
\end{aligned}
\end{equation}
% \{Loss functions using by Other methods: \}
where {$score_{1}$ and $score_{2}$ are STS scores used to measure the similarity between two pieces of text, 
with scores typically ranging from 0 (no similarity) to 1 (identical meaning). In previous methods, language models were trained to predict STS scores between text containing IEs and those without IEs, in order to improve their ability to understand IEs.

In our method, we will utilize a triplet loss in combination with a miner to extract triplets without STS scores, approximating the definition in equations (\ref{eq0}).  It is worth noting that without using STS scores, training data can be acquired more easily.

% \subsection{Triplet Loss Employment}
% \label{sec:Triplet}

\subsection{Mining to Extract Triplets}
The original dataset only has IEs, the sentences with IE and their correct and incorrect paraphrases.
% In this section, we will explain how 
To extract triplets for our idiomatic-aware training we use a semantic meaning miner.We use batch negatives approach that leverages the other samples present in the same mini-batch for serving as negative instances. However, not all negatives in a batch are useful for our training. Thus, we introduce a special preprocessing step in our method.
%Labels will be assigned based on semantic meaning; sentences with the same meaning have the same label see Table \ref{tab:examples. In this way, $anchor$, $positive$, and $negative$ can be distinguished.  }

\begin{table*}[h!]
\small
\centering
\begin{tabularx}{\textwidth}{|c|c|c|X|}
    \hline
    \textbf{Group} &\textbf{Index} &\textbf{Label} &\textbf{Instance} \\
    \hline
    \multirow{3}{4em}{$1$} & 1&\cellcolor{gray!30}en1& \cellcolor{gray!30}So Aaron faced the same brutal racism other Black players of the era experienced, especially as the slugger approached Ruth's  \textbf{IDhomerunID}  record. \\
    \cline{2-4}
     & 2&\cellcolor{gray!30}en1 &\cellcolor{gray!30}So Aaron faced the same brutal racism other Black players of the era experienced, especially as the slugger approached Ruth's \textbf{baseball run} record.\\
    \cline{2-4}
     & 3&\textit{en2} &  So Aaron faced the same brutal racism other Black players of the era experienced, especially as the slugger approached Ruth's \textbf{house run} record.\\
    
         \hline
        \hline
    \multirow{4}{4em}{$2$} & 4&\cellcolor{gray!30}en3&\cellcolor{gray!30}Robinhood is supposed to be the revolutionary trading app that made it possible for the \;\;\textbf{IDsmallfryID}\;\; to get together and crush the big boys.  \\
    \cline{2-4}
     & 5&\cellcolor{gray!30}en3 &\cellcolor{gray!30}Robinhood is supposed to be the revolutionary trading app that made it possible for the \;\; \textbf{insignificant} \;\; to get together and crush the big boys.  \\
    \cline{2-4}
    & 6&\textit{en4} &Robinhood is supposed to be the revolutionary trading app that made it possible for the \;\; \textbf{little fry} \;\; to get together and crush the big boys. \\
    \cline{2-4}
    & 7&\textit{en5} &Robinhood is supposed to be the revolutionary trading app that made it possible for the \;\; \textbf{little kid} \;\; to get together and crush the big boys.\\
    \hline
\end{tabularx}
\caption{Examples of training data. Sentences that have the same meaning are given the same labels.  We treat IE expressions as a single token and preprocess it as shown. For example, the IE \textit{home run} is replaced as \textit{IDhomerunID}.}
\label{tab:examples}
\end{table*}

\paragraph{Relabel Training Data} 
% [IF IT IS A PROCESS OF RELABELING (CHANGING A LABEL) WHAT CRITERIA WAS USED TO ASSIGN THE INITIAL LABELS?]
For a triplet to be valid, it must meet certain requirements. We first categorize sentences into different groups. Each group contains a sentence with the IE ($s$), its correct ($c$) and incorrect ($i$) paraphrases, such as examples in Table \ref{tab:examples}. New labels will be assigned in each group based on IEs and their paraphrases. Our approach assigns identical labels to sentences with the same meaning (original sentence and correct paraphrases).
Firstly, $s$ and $c$ must have the same label, which means they represent the same meaning. Secondly, each $i$ must have different labels and differ from the label of $s$ and $c$, which means they represent different meanings. 
% By comparing the feature vectors of $a$, $p$, and $n$, the language modeling system can determine if $a$ and $p$ belong to the same meaning or not.

% New labels will be assigned based on IEs and their paraphrases; 
% sentences with the IE and correct paraphrases will have the same label,  and others will have different labels see Table \ref{tab:examples}.
It also needs labels in different groups to be distinct to others. For example in Table \ref{tab:examples}, as sentences with index 4 and 5 are a pair of $s$ and $c$, they are assigned with the same label {\textbf{en3}}. Other sentences in Group 2 are assigned different labels because they are incorrect paraphrases. The labels in Group 2 are distinct from labels in Group 1.

In this way,  a triplet can be acquired easily since $anchor$, $positive$ are sentences with the same labels, and a $negative$ is a sentence with different labels.
\paragraph{Selected Multi-negatives}
% To differentiate between correct and incorrect paraphrases of idiomatic expressions, we create specific triplets based on sentences with semantic meanings. 
Negative instances refer to sentences whose labels differ from the anchor and positive in a batch.
In the case of Multi Negative Ranking Loss \citep{sun2020circle} with triplet formation, there are multiple negatives $[n_1, n_2, ...n_k]$ for each anchor-positive pair, and the objective is to ensure that the anchor is closer to the positive than to any of the negatives by a margin. 
\begin{equation} \label{eq:multiN}
\begin{split}
&\mathcal{L}_{\text{multi-negative}}(a, p, [n_1, \ldots, n_k])\\
&= \sum_{i=1}^{k} \max( d(a, p) - d(a, n_i) + m,0)
\end{split}
\end{equation}
We take the SemEval 2022 task 2B training set as our source to build our training data. The dataset comprises approximately $8,600$ annotated examples in multiple languages, including English and Portuguese. It was divided into 4,840 training sentences, 739 development sentences, 483 evaluation sentences and 2,342 test sentences. The original training data already includes information on correct and incorrect paraphrases. The context sentences help disambiguate the IE's meaning. This annotated data is crucial for training machine learning models to detect idiomatic expressions in varied linguistic contexts, facilitating multilingual natural language understanding and processing.

After relabeling, the training dataset will be a list of sentences with their corresponding label. We do not shuffle the training data to maintain its order, as sentences that belong to a triplet are adjacent in the training set. The batch size is set to 64, which is a balance between easy training and ensuring a sufficient number of sentences to build triplets. 

However, not all negatives contribute equally to our learning. Some triplets may already satisfy the constraint (easy triplets), such as triplets with negatives that are sentences from other groups. They provide little to no information of IEs understanding for the model to learn from.
\paragraph{Mine Triplets}
In our methods, a semantic similarity miner calculates Euclidean distance between all possible pairs of embeddings in a batch and selects according to its similarity margin. The miner similarity margin is the difference between the anchor-positive distance and the anchor-negative distance. It is also a hyperparameter in our method. 
The miner select the triplets that violate the miner similarity margin to make the model learn nuanced differences between figurative and literal meanings of IEs. 
% , ensuring that the triplet loss only focuses on sentences with similar semantics. 
For instance, a triplet of sentences could include an idiomatic expression as the anchor, its paraphrases as the positive, and a sentence with a literal meaning as the negative. 

% While training, previous methods consider these negative sentences in a batch[IS THIS BATCH THE SAME BATCH MENTIONED BEFORE?], as described in \citep{gao-etal-2021-simcse}. 

% By following this strategy, the sentences $S_{MWE}$ and $S_{\rightarrow c}$ form positive pairs $(a_i, p_i)$, while all other ${S_{\rightarrow c}}$s in the batch are considered negatives. 

% To build our triplets, by setting an anchor, a sentence with the same label as the anchor's label is set to positive, and a sentence with a different label [ HOW A LABEL IS ACQUIRED? ] is set to negative. Sentences with the same label could be the anchor and the positive of a triplet.  A sentence with a different label will be the negative.

 % We first relabel the data that $S_{MWE}$ and $S_{\rightarrow c}$ with the same label and sentences other than them with different labels. The positives of each anchor are predefined.
Table \ref{tab:examples} illustrates the newly build training data. In this case, $S_{MWE}$ and $S_{\rightarrow c}$ can act as anchor and positive to each other, and ${S_{\rightarrow i}}$ can only be treated as the negative in a triplet. It is worth noting that $S_{MWE}$ and $S_{\rightarrow c}$ are interchangeable to form pairs $(a_i, p_i)$, which can build more triplets for our training.
For example, in Table \ref{tab:examples}, with the miner, it will only take sentences in the same group because the semantic meanings of different groups are not similar. In this way, sentences in Group 1 can build 2 triplets with index 1 and 2 being the anchor and positive interchangeably. Sentences in Group 2 can build 4 triplets.

% \begin{align}
%  \label{eq6} 
%     \small{
%    sim(a, p) - sim(a, n) = m}
% \end{align}
\subsection{Objective Transformation}
In our approach, both $S_{MWE}$ and $S_{\rightarrow c}$ can serve as anchors. However, since we assign different labels to various incorrect paraphrases, no positive sentence in a group can be associated with any $S_{\rightarrow i}$ as the anchor. As a result, there are only two possible scenarios in our approach.

If $S_{MWE}$ is the anchor,
\begin{equation} \label{eq7}
% \small
   sim(S_{MWE}, S_{\rightarrow c}) - sim(S_{MWE}, S_{\rightarrow i}) \le m_a
\end{equation}

\noindent if $S_{\rightarrow c}$ is the anchor,
\begin{equation} \label{eq8}
   sim(S_{\rightarrow c}, S_{MWE} ) - sim(S_{\rightarrow c}, S_{\rightarrow i}) \le m_b
\end{equation}
The margin $m$ is a predefined fixing value. If we set $m_a$ = $m_b$, then combining Eq. (\ref{eq7}) and  Eq. (\ref{eq8}), the objective function can be transformed to:
\begin{equation} \label{eq9}
\begin{split}
   sim(S_{MWE}, S_{\rightarrow c}) - sim(S_{MWE}, S_{\rightarrow i}) \approx \\sim(S_{\rightarrow c}, S_{MWE} ) - sim(S_{\rightarrow c}, S_{\rightarrow i})
\end{split}
\end{equation}
The similarity measure is symmetric, therefore $sim(S_{MWE}, S_{\rightarrow c}) = sim(S_{\rightarrow c}, S_{MWE} )$.} In this way, our objective function equivalent to:
\begin{align} \label{eq10}
    sim(S_{MWE}, S_{\rightarrow i}) \approx sim(S_{\rightarrow c}, S_{\rightarrow i}) 
\end{align}
Equation (\ref{eq10}) approximates the definition of the good idiomatic aware model in Equation (\ref{eq0}). In this way, by using our specific triplet loss, we can train a model to be idiomatically aware more directly without STS scores. 

% \begin{align} \label{eq11}
%     sim(S) = sim(S_{\rightarrow c}) 
% \end{align}

% In this situation, 

% \subsection{Data augmentation with GPT}

\section{Experiment}
\label{sec:Ex}
This section presents the comprehensive methodology employed to our model training. We begin by detailing the experiment implementation, including the hyperparameter setting, models used, evaluation method, and the overall setup. 
% Through this approach, we aim to provide a thorough account of the experimental process, ensuring transparency and reproducibility of our findings.
\subsection{Implementation Details}

The method is implemented by using the Transformers \citep{wolf2020transformers} and PyTorch Metric Learning \citep{Musgrave2020PyTorchML} libraries. Some of the pre-trained models are fetched from Sentence-transformer library\footnote{https://www.sbert.net/docs/pretrained\_models.html} and HuggingFace Model repositories\footnote{https://huggingface.co/models}. 

We calculate sentence similarity using the cosine similarity of the mean pooling of the last two hidden layers. Empirically, we set the similarity margin for the miner to $0.4$, and the training loss margin to $0.3$. Given the limited availability of idiomatic text data, relying solely on the training signal from our contrastive objective is insufficient for learning general semantic representations. Therefore, we initialize our model with other pre-trained semantic-aware models \citep{reimers-gurevych-2019-sentence}.  
Our best model uses a pre-trained multilingual model, ‘\textit{paraphrase-multilingual-mpnet-base-v2}’\footnote{https://huggingface.co/sentence-transformers/paraphrase-multilingual-mpnet-base-v2}, and fine-tunes it with our method to fit the task.
It is pre-trained with millions of paraphrases, so it can represent sentence semantic meanings well \citep{reimers-gurevych-2019-sentence}.

\subsection{Evaluation}
We perform intrinsic evaluation \citep{reimers-etal-2016-task} using the SemEval-2022 task 2 Subtask B\footnote{https://codalab.lisn.upsaclay.fr/competitions/8121} (Sem2B). We use Spearman's rank correlation ($\rho$) between model-generated scores and human judgment scores to see how well models understand idioms in sentences. Instead of comparing exact scores, this method focuses on how the sentence pairs are ranked based on predicted similarity compared to human judgments. A higher correlation means the model is better at understanding relationships, including those involving idioms, even if the exact predicted scores themselves aren't always perfect matches.
% This task is significant because it is the first shared task to focus on both idiomaticity detection and representation. 
% Idiomaticity is a complex phenomenon that can be manifested in different ways, for example, an idiomatic expression may have a non-literal meaning, or it may have a different meaning depending on the context in which it is used.

% We also take the NCTTI \citep{garcia2021probing} dataset to assess the ability of vector models to capture idiomaticity. It contains annotations for idiomaticity at both the type and token levels for 280 noun compounds in English and 180 noun compounds in Portuguese.  
\subsection{Comparative Analysis}

We compare our method with well-performed Semantic Textual Similarity modelsand recent large language models (LLMs). Some training-based methods are from SemEval-2022 task 2 Fine Tune solutions \citep{madabushi2022semeval}. 
Here are brief descriptions:
\begin{description}
\item[YNU-HPCC] \citep{liu-etal-2022-ynu} is the previous best method, which uses contrastive learning approaches in sentence representation. However, it treats negatives in a batch equally. 

\item[drsphelps] \citep{phelps2022drsphelps} introduces a method for improving idiom representation in language models by incorporating idiom-specific embeddings using BERTRAM into a BERT sentence transformer.

\item[baseline] is the SemEval task's baseline results. It is fine-tuned using multilingual BERT \citep{devlin-etal-2019-bert} and adding single tokens for each MWE in the data.

\item[GTE large]\footnote{https://huggingface.co/thenlper/gte-large} is a powerful text embedding model trained with multi-stage contrastive learning, delivers impressive performance across NLP and code tasks despite its modest size \citep{li2023towards}.

\item[E5 large]\footnote{https://huggingface.co/intfloat/e5-large} uses weakly-supervised contrastive pre-training for text embeddings that achieves excellent for general-purpose text representation \citep{wang2022text}.
\item{\textbf{LLama2}} \citep{touvron2023llama} achieved excellent performance in a series of NLP tasks.
We select the \textit{LLama2-13B} for comparison.
\end{description}

\section{Results and Analysis}
\label{sec:Re}
In this section, we report results and analyze them in different settings. 
\subsection{Overall Results}
\begin{table}
\centering
\small
\begin{tabular}{ |c|c|c|c|c|c| } 
\hline
\multirow{2}{*}{Method}& \multirow{2}{*}{Model Size}& \multicolumn{2}{c|}{Subset}  & \multirow{2}{*}{All} \\
\cline{3-4}
&&Idiom  & STS &\\
\hline
YNU-HPCC&183M& 0.428 &0.664&0.665 \\ 
drsphelps&420M& 0.412 &\textbf{0.819}& 0.650\\ 

baseline& 110M& 0.399 &0.596& 0.595 \\
\hline
GTE large&334M & 0.236  & 0.806& 0.465 \\
E5 large&334M& 0.252 & 0.807& 0.514 \\ 

LLama2&13,000M& 0.171 &  0.486& 0.399 \\
\hline
Our best&558M& \textbf{0.548} &0.716& \textbf{0.690} \\
\hline

\end{tabular}
\caption{ Test results of Task 2 on Spearman’s rank correlation
coefficient between the two sets of STS scores. }
\label{tab:all}
\end{table}

Table \ref{tab:all} demonstrates that our method outperforms all other models both overall and in the Idiom Only subset. 
The "STS only" score refers to the performance of systems on Semantic Text Similarity data that does not necessarily contain idioms. In contrast, the "Idiom only" score pertains to the performance on idiom STS data. "All" represents the overall performance of a model across the entire dataset.
In the Idiom Only subset, our method achieves a score of $0.548$, which is higher than the score of the next best model, YNU-HPCC ($0.428$). In the overall performance, it achieves a score of $0.690$, exceeding the score of the next best model, YNU-HPCC ($0.665$). In the STS subset, drsphelps achieves the highest score of $0.819$. These results suggest that our method is a powerful and effective idiom-aware text embedding model that can be used for a variety of idiomatic expressions related NLP tasks.

GTE large and E5 large both show a similar pattern of lower performance in the Idiom task ($0.236$ and $0.252$, respectively) but strong results in the STS task ($0.806$ and $0.807$, respectively). Their overall scores ($0.465$ and $0.514$, respectively) suggest that while they are proficient in semantic textual similarity, their capacity to handle idiomatic expressions is not as developed.
LLama2 has the lowest scores across all three categories, with a particularly low score for Idiom ($0.171$). It reveals a surprising lack of ability to represent idiomatic expressions for such recent general large language model. 
% It is important to note that the results in this table are based on a single experiment, and more research is needed to confirm these findings. Additionally, the table does not show the performance of other text embedding models that may be relevant.
\subsection{Performance on Different Languages}
\begin{table}
\centering
\begin{tabular}{ |c|c|c|c|c| } 
\hline
\multirow{2}{*}{Language} & \multicolumn{2}{c|}{Subset}  & \multirow{2}{*}{All} \\
\cline{2-3}
&Idiom Only & STS Only&\\
\hline
EN& 0.560 &  \bf{0.759}& \bf{0.757} \\ 
PT&  \bf{0.570} & 0.657& 0.707\\ 
GL& 0.515 & -& 0.515 \\ 
3L& 0.548 & 0.716& 0.690 \\ 
\hline

\end{tabular}
\caption{ Our test results of Task 2 on Spearman’s rank correlation coefficient in English (EN), Portuguese (PT), and Galician (GL) separately. 3L is the combination of 3 languages.}
\label{tab:language}
\end{table}

\begin{table}[h!]
    \centering
    \begin{tabular}{|l|c|c|c|c|}
        \hline
        \textbf{Method} & \textbf{Lang(s)} & \textbf{Idiom} & \textbf{STS} & \textbf{ALL} \\ \hline
        \multirow{4}{*}{drsphelps} & EN & \textbf{0.486} & 0.834 & \textbf{0.764} \\ 
        & PT & 0.464 & \textbf{0.791} & 0.731 \\ 
        & GL & \textbf{0.286} & - & 0.286 \\ 
        & 3L & 0.412 & 0.819 & 0.650 \\ \hline
        \multirow{4}{*}{E5 large} & EN & 0.242 & \textbf{0.919} & 0.607 \\ 
        & PT & 0.276 & 0.646 & 0.551 \\ 
        & GL & 0.247 & - & 0.247 \\ 
        &3L & 0.252 & 0.807 & 0.514 \\ \hline
        \multirow{4}{*}{LLama2} & EN & 0.156 & 0.512 & 0.391 \\ 
        & PT & 0.206 & 0.496 & 0.510 \\ 
        & GL & 0.185 & - & 0.185 \\ 
        & 3L & 0.171 & 0.486 & 0.399 \\ \hline
    \end{tabular}
    \caption{Spearman's rank correlation coefficients for drsphelps, E5 large, and LLama2 methods across idiomatic only, STS only, and overall results in English (EN), Portuguese (PT), Galician (GL), and their combination (3L). }
    \label{tab:language2}
\end{table}

The results in Table \ref{tab:language} show that our best model performed well on Sem2B and in all three languages. The best results were achieved, with overall $\rho$ values of $0.757$ for English, $0.707$ for Portuguese, and $0.515$ for Galician. The best overall results on Sem2B were achieved for English, and the best Idiom Only score was achieved for Portuguese. There is no STS-only score for Galician in the test set.  The models performed best on English, followed by Portuguese and Galician. This is due to the fact that there is more training data available for English than for Portuguese or Galician. The results also show that the models were able to generalize well, even when the amount of training data was limited. For example, the models achieved $\rho$ values of $0.707$ and $0.515$ for Portuguese and Galician, even though the training data for these two languages was smaller than the training data for English. Compared to other methods in Table \ref{tab:language2}, our model excels particularly in handling idiomatic expressions, outperforming other models in the Idiom Only subset. Additionally, while drsphelps and E5 large show strong results in the STS subset, our model maintains a balanced performance across all datasets, demonstrating its robustness.

\subsection{Impact of Our Training}
\begin{table}
\centering
\begin{tabular}{ |c|c|c|c|c| } 
\hline
\multirow{2}{*}{Model} & \multicolumn{2}{c|}{Subset}  & \multirow{2}{*}{All} \\
\cline{2-3}
&Idiom Only & STS Only&\\
\hline
\multicolumn{4}{|c|}{Original}\\
\hline
roberta-base& 0.184 & 0.626& 0.492 \\ 
x-r-large& 0.138 & 0.284& 0.444 \\  
p-v2& 0.225 & 0.838& 0.532\\ 
\hline
\multicolumn{4}{|c|}{After Training}\\
\hline
roberta-base& 0.454 & 0.622& 0.613 \\ 
x-r-large& 0.484 & 0.465&0.639 \\
p-v2& 0.548 & 0.716& 0.690 \\ 

\hline

\end{tabular}
\caption{ Test results across three models \textit{roberta-base}, \textit{xlm-roberta-large} (x-r-large) and \textit{paraphrase-multilingual-mpnet-base-v2} (p-v2)
before and after training. }
\label{tab:after}
\end{table}
Table \ref{tab:after} presents comparative performance results of three language models, roberta-base \citep{liu2019roberta}, xlm-roberta-large \citep{conneau-etal-2020-unsupervised} (x-r-large) and \emph{paraphrase-multilingual-mpnet-base-v2} (p-v2), across three different subsets of data: Idiom Only, STS Only, and All. The first two models are widely used language models with general and multilingual properties, respectively. The third model is the base model we used in our best model. The results are split into two categories: `Original', which indicates the performance before additional training, and `After Training', showing the performance post-training.

For the Idiom Only subset, the original scores were $0.184$ for roberta-base,  $0.138$ for x-r-large,  and $0.225$ for p-v2. After training, these scores improved significantly to $0.454$ for roberta-base, $0.484$ for x-r-large and $0.548$ for p-v2. When looking at the overall performance,  the x-r-large model's performance originally was $0.444$ and increased to $0.639$ after training. Similarly, the p-v2 model's performance was initially $0.532$ and rose to $0.690$ after training. In the STS Only subset, there have been declines at p-v2 from $0.838$ to $0.716$. It is because our training only focuses on improving idiom representation, and it may slightly sacrifice the performance of specific fully-trained models. 

The size of our model's parameters is slightly larger than most, but it significantly outperforms others, demonstrating the effectiveness of our proposed method beyond just using a larger model. As shown in Table 4, our method achieves superior results in idiomatic representation even when compared with implementations using the same model sizes.

The results in Table \ref{tab:epoch} showcase that as the number of epochs increases, the overall performance as well as the performance on the "Idiom Only" subset generally improves. This suggests that the model is learning and improving its IE understanding ability during our training.
The performance on the "Idiom Only" subset starts very low at epoch 0, with an accuracy of $0.225$, which is expected since the model has not learned much IE representation yet. There is a significant improvement between epoch 0 and epoch 8, with the score nearly doubling to 0.499. The improvement in performance starts to plateau after epoch 10, with only minor increases observed at epochs 15 and 25.
The "STS Only" subset starts with a high performance even at epoch 0, with an accuracy of $0.838$. This is because the model has already been pre-trained with STS tasks. Unlike the "Idiom Only" subset, the performance on the "STS Only" subset decreases as the number of epochs increases, dropping to $0.716$ by epoch 25. This could indicate that the model is becoming more specialized in the idiom task at the expense of the STS task. In summary, while the model is improving in its ability to understand idioms with more training, this comes at the cost of its performance on STS tasks. This trade-off can be addressed by adjusting the training process.

In summary, our models were able to generalize well to different settings, even when the amount of training data was limited. This suggests that the models are learning to capture the underlying properties of idiomatic expressions, rather than simply memorizing a list of idiomatic expressions.

% \begin{table}
% \centering
% \begin{tabular}{ |c|c|c| } 
% \hline
% \multirow{2}{*}{Model} & \multicolumn{2}{c|}{Subset}\\  
% \cline{2-3}
% & Modifier only & Head only\\
% \hline
% p-v2 & 0.195&0.178 \\
% roberta-large &0.203 &0.144\\
% roberta-base & 0.154 & 0.103 \\
% XLM-Roberta-base& 0.144 & 0.017 \\ 
% bert-base& 0.185 & 0.067 \\ 
% % Zhichun Road& 0.3956 & 0.5615& 0.6401 \\ 
% % baseline& 0.3990 & 0.5961& 0.5951 \\
% Our model& \textbf{0.380} & \textbf{0.281} \\
% \hline

% \end{tabular}
% \caption{ Training Results on NCTTI with different models.?? }
% \end{table}

\begin{table}
\centering
\begin{tabular}{ |c|c|c|c| } 
\hline
\multirow{2}{*}{Epoch} & \multicolumn{2}{c|}{Subset}& \multirow{2}{*}{All}\\  
\cline{2-3}
& Idiom Only & STS Only&\\
\hline
0 & 0.225& 0.838&0.532\\
8 & 0.499&0.785& 0.670\\
10 &0.531 &0.740&0.682\\
15 & 0.539 & 0.740& 0.688\\
25& 0.548 & 0.716& 0.690\\ 
% bert-base& 0.185 & 0.067& \\ 
% Zhichun Road& 0.3956 & 0.5615& 0.6401 \\ 
% baseline& 0.3990 & 0.5961& 0.5951 \\
% Our model& \textbf{0.380} & \textbf{0.281}& \\
\hline

\end{tabular}
\caption{ Test Results with different training epochs by using same \textit{p-v2} model.  }
\label{tab:epoch}
\end{table}
\section{Discussion}

% For reducing the training difficulty, the idiomatic expressions in each sentence are identified beforehand. It means that the proposed model could perform worse with the text without IE identified.
The proposed model for training requires the identification of idiomatic expressions (IEs) in each sentence beforehand. This step is crucial for reducing the difficulty of the training process. Without identifying the IEs beforehand, the model may not perform optimally, and its accuracy may be compromised. Therefore, it is essential to ensure that the text has IEs identified to achieve the best results.

\section{Conclusion}
Idiom representations have always been a challenge due to the non-compositional nature of idiomatic expressions. The performance of downstream tasks, such as translation and simplification, is dependent on the quality of the representations. This paper proposes a new method to train language models using adaptive contrastive learning with triplets and resampling miners. In this way, our method can build a better optimization objective, which makes the training very efficient.

The proposed method, evaluated on the idiomatic semantic text similarity tasks, significantly outperforms previous methods. With limited idiomatic text data, the sole training signal of the contrastive objective is not sufficient to learn general semantic representations. Therefore, the model is initialized with other pre-trained semantic-aware models. A series of base models in different sizes and pre-training strategies are trained in the proposed training loss. The best models achieve new state-of-the-art results with a significant improvement in overall over the previous best in the evaluation task.

\section{Future Work}

In the future, we plan to use the idiomatic-aware model in other NLP tasks that require sensitivity to idiomatic expressions, such as machine translation. Additionally, we aim to improve the model's training by adding more supervision, which will help it focus on contextual information. This will allow the model to better understand multiword expressions based on different contexts.
\section{Limitations}
In order to train our model, we require triplets that consist of three distinct parts: a sentence that contains IEs, a correct paraphrase of those IEs, and an incorrect paraphrase of those IEs. The quality of triplets is crucial to the development of our model and requires intensive human expert involvement to ensure accuracy.
\section{Acknowledgments}
The work was partly supported by EPSRC grant EP/T02450X/1,  by  NAF/R2/202209 grant and by Brazilian CNPq Grant 311497/2021-7.

\bibliography{anthology,custom}

\appendix

%\section{Example Appendix}
%\label{sec:appendix}
% 1.using parts of the MWEs(measuring similarities of pairs with each part only )
% 2. CDT conference
% 3. Multi-model conference
% 4. Darwin team data
% 5. other methods

\end{document}